\documentclass[11pt,a4paper]{article}

\usepackage[hyperref]{acl2021}
\usepackage{times}
\usepackage{latexsym}

\usepackage{pifont}
\newcommand{\xmark}{\ding{55}}%

\usepackage{microtype}
\usepackage{mathtools}

\aclfinalcopy 

\usepackage{paralist}
\usepackage{anyfontsize}

\usepackage{xspace}

\usepackage{amsmath}
\usepackage{amssymb}
\usepackage{graphicx}
\usepackage{multirow}
\usepackage{booktabs}
\usepackage{multirow}
\usepackage{array}
\usepackage{diagbox}

\newcommand{\KB}{PAQ}

\newcommand{\QAEngine}{RePAQ}

\usepackage[acronym]{glossaries}

\usepackage[capitalise]{cleveref}
\newacronym{odqa}{ODQA}{Open-Domain question Answering}
\newacronym{kb}{KB}{Knowledge Base}
\newacronym{kbqa}{KB-QA}{Question Answering over Knowledge Bases}
\title{\KB{}: 65 Million Probably-Asked Questions and \\ What You Can Do With Them}

\author{
  Patrick Lewis$^{\dagger{}\ddagger{}}$ \hspace{0.15cm} 
  Yuxiang Wu$^{\ddagger{}}$ \hspace{0.15cm} 
  Linqing Liu$^{\ddagger{}}$ \hspace{0.15cm}
  Pasquale Minervini$^{\ddagger{}}$ \hspace{0.15cm}
  Heinrich K\"uttler$^{\dagger{}}$ \hspace{0.15cm}\\
  \textbf{Aleksandra Piktus$^{\dagger{}}$ \hspace{0.15cm}
  Pontus Stenetorp$^{\ddagger{}}$ \hspace{0.15cm}
  Sebastian Riedel$^{\dagger{}\ddagger{}}$ \hspace{0.15cm}} \\
  $^{\dagger{}}$Facebook AI Research \hspace{0.3cm} $^{\ddagger{}}$University College London \\
  \texttt{plewis@fb.com}
}

\date{}

\begin{document}
\maketitle
\begin{abstract}

Open-domain Question Answering models which directly leverage question-answer~(QA) pairs, such as closed-book QA~(CBQA) models and QA-pair retrievers, show promise in terms of speed and memory compared to conventional models which retrieve and read from text corpora.
QA-pair retrievers also offer interpretable answers, a high degree of control, and are trivial to update at test time with new knowledge.
However, these models lack the accuracy of retrieve-and-read systems, as substantially less knowledge is covered by the available QA-pairs relative to text corpora like Wikipedia.
To facilitate improved QA-pair models, we introduce \emph{Probably Asked Questions} (\KB{}), a very large resource of 65M automatically-generated 
QA-pairs.
We introduce a new QA-pair retriever, \QAEngine{}, to complement \KB{}. 
We find that \KB{} \emph{preempts} and \emph{caches} test questions, enabling \QAEngine{} to match the accuracy of recent retrieve-and-read models, whilst being significantly faster.
Using \KB{}, we train CBQA models which outperform comparable baselines by 5\%, but trail \QAEngine{} by over 15\%, indicating the effectiveness of explicit retrieval. 
\QAEngine{} can be configured for size (under 500MB) or speed (over 1K questions per second) whilst retaining high accuracy.
Lastly, we demonstrate \QAEngine{}'s strength at \emph{selective QA}, abstaining from answering when it is likely to be incorrect. 
This enables \QAEngine{} to ``back-off" to a more expensive state-of-the-art model,
leading to a combined system which is both more accurate and 2x faster than the state-of-the-art model alone.

\end{abstract}

\section{Introduction}
Open-domain QA~(\acrshort{odqa}) systems usually have access to a background corpus that can be used to answer questions.
Models which explicitly exploit this corpus are commonly referred to as \emph{Open-book} models~\cite{roberts_how_2020}. 
They typically index the whole corpus, and then \emph{retrieve-and-read} documents in order to answer questions on-the-fly~\cite[inter alia]{chen_reading_2017,lee_latent_2019}.

A second class of models, \emph{closed-book} question answering (CBQA) models, have recently been proposed. 
They learn to directly map questions to answers from training question-answer (QA) pairs without access to a background corpus~\cite{roberts_how_2020,ye_studying_2021}. 
These models usually take the form of pretrained seq2seq models such as T5~\cite{raffel_exploring_2020} or BART~\cite{lewis_bart_2019}, fine-tuned on QA-pairs. 
It has recently been shown that current closed-book models mostly memorise training QA-pairs, and can struggle to answer questions that do not overlap with training data~\cite{lewis_question_2020}.

Models which explicitly retrieve (training) QA-pairs, rather than memorizing them in parameters, have been shown to perform competitively with CBQA models~\cite{lewis_question_2020,xiao_open-domain_2020}.
These models have a number of useful properties, such as fast inference, interpretable outputs (by inspecting retrieved QA-pairs), and the ability to update the model's knowledge at test time by adding or removing QA-pairs.

However, CBQA and QA-pair retriever models are currently not competitive with retrieve-and-read systems in terms of accuracy, largely because the training QA-pairs they operate on cover substantially less knowledge than background corpora like Wikipedia.
In this paper, we explore whether massively expanding the coverage of QA-pairs enables CBQA and QA-pair retriever models which are competitive with retrieve-and-read models.

We present Probably Asked Questions~(\KB{}), a semi-structured Knowledge Base~(KB) of 65M natural language QA-pairs, which models can memorise and/or learn to retrieve from.
\KB{} differs from traditional KBs in that questions and answers are stored in natural language, and that questions are generated such that they are likely to appear in ODQA datasets.
\KB{} is automatically constructed using a question generation model and Wikipedia.
To ensure generated questions are not \emph{only} answerable given the passage they are generated from, we employ a \emph{global filtering} post-processing step employing a 
state-of-the-art ODQA system. 
This greatly reduces the amount of wrong and ambiguous questions compared other approaches~\cite{fang_accelerating_2020,alberti_synthetic_2019}, and is critical for high-accuracy, downstream QA models.

To complement \KB{} we develop \QAEngine{}, a question answering model based on question retrieval/matching models, using dense Maximum Inner Product Search-based retrieval, and optionally, re-ranking.
We show that \KB{} and \QAEngine{} provide accurate ODQA predictions, at the level of
relatively recent large-scale retrieve-and-read systems such as RAG~\cite{lewis_retrieval-augmented_2020} on  NaturalQuestions~\cite{kwiatkowski_natural_2019} and TriviaQA~\cite{joshi_triviaqa:_2017}.
\KB{} instances are annotated with scores that reflect how likely we expect questions to appear, which can be used to control the memory footprint of \QAEngine{} by filtering the KB accordingly.
As a result, \QAEngine{} is extremely flexible, allowing us to configure QA systems with near state-of-the-art results, very small memory size, or inference speeds of over 1,000 questions per second.
Memory-optimised configurations of \QAEngine{} won two of the four tracks of the 2020 EfficientQA NeurIPS competition~\citep{min_neurips_2020}, with system sizes of 336MB and 29MB, respectively.

We also show that \KB{} is a useful source of training data for CBQA models. 
BART models trained on \KB{} outperform baselines trained on standard data by 5\%.
However, these models struggle to effectively memorise all the knowledge in \KB{}, lagging behind \QAEngine{} by 15\%.
This demonstrates the effectiveness of \QAEngine{} at leveraging \KB{}. 

Finally, we show that since \QAEngine{}'s question matching score correlates well with QA accuracy, it effectively ``knows when it doesn't know'', allowing for \emph{selective question answering}~\cite{rodriguez_quizbowl_2019} where QA systems may abstain from answering if  confidence is too low. 
Whilst answer abstaining is important in its own right, it also enables an elegant ``back-off'' approach where we can defer to a more accurate but expensive QA system when answer confidence is low.
This enables us to make use of the best of both speed and accuracy.

In summary, we make the following contributions:
\begin{inparaenum}[i)]
    \item{introduce \KB{}, 65M QA-pairs automatically generated from Wikipedia, and demonstrate the importance of global filtering for high quality}
    \item{introduce \QAEngine{}, a QA system designed to utilize \KB{} and demonstrate how it can be optimised for memory,  speed or accuracy}
    \item{investigate the utility of \KB{} for CBQA models, improving by 5\% but note significant headroom  to \QAEngine{}}
    \item{demonstrate \QAEngine{}'s strength on selective QA, enabling us to combine \QAEngine{} with a state-of-the-art QA model, making it both more accurate and 2x faster\footnote{The \KB{} data, models and code will be made available at \url{https://github.com/facebookresearch/PAQ}}
}
\end{inparaenum}
\section{Open-Domain Question Answering}

ODQA is the task of answering natural language factoid question from an open set of domains. 
A typical question might be ``when was the last year astronauts landed on the moon?", with a target answer ``1972''.
The goal of ODQA is to develop an answer function $m : Q \mapsto A$, where $Q$ and $A$ respectively are the sets of all possible questions and answers.
We assume there is a distribution $P(q,a)$ of QA-pairs, defined over $Q \times A$.
A good answer function will minimise the expected error over $P(q,a)$ with respect to some loss function, such as answer string match.
In practice, we do not have access to $P(q,a)$, and instead rely on an empirical sample of QA-pairs $\mathcal{K}$ drawn from $P$,
and measure the empirical loss of answer functions on $\mathcal{K}$.
Our goal in this work is to implicitly model $P(q,a)$ so that we can draw a large sample of QA-pairs, \KB{}, which we can train on and/or retrieve from. Drawing a sufficiently large sample will overlap with $\mathcal{K}$, essentially \emph{pre-empting} and \emph{caching} questions that humans may ask at test-time.
This allows us to shift computation from test-time to train-time compared to retrieve-and-read methods.
\section{Generating Question-Answer Pairs}
\label{sec:generating_question_answer_pairs}
\begin{figure*}[t]
\centering
\hskip-2mm
  \includegraphics[width=.95\textwidth]{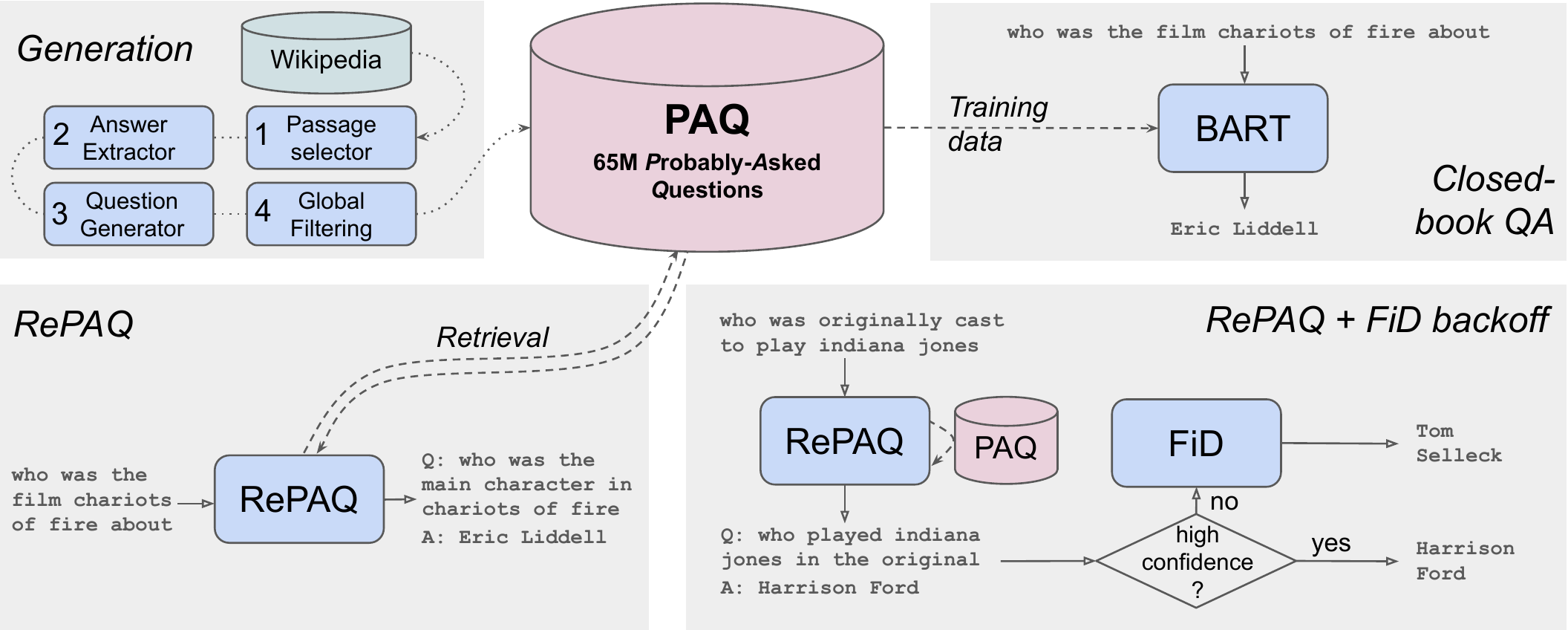}

  \caption{
  Top Left: Generation pipeline for QA-pairs in \KB{}. 
  Top Right: \KB{} used as training data for CBQA models.
  Bottom Left: \QAEngine{} retrieves similar QA-pairs to input questions from \KB{}. 
  Bottom right: \QAEngine{}'s confidence is predictive of accuracy. If confidence is low, we can defer to slower, more accurate systems, like FiD. 
    } 
  \label{fig:generation_pipeline}
\end{figure*}
In this section, we describe the process for generating \KB{}.
Given a large background corpus $C$, our QA-pair generation process consists of the following components:
\begin{enumerate}
\itemsep0em 
\item A \emph{passage selection} model $p_{s}(c)$, to identify passages which humans are likely to ask questions about.
\item An \emph{answer extraction} model $p_{a}(a \mid c)$, for identifying spans in a passage that are more likely to be answers to a question.
\item A \emph{question generator} model $p_{q}(q \mid a, c)$ that, given a passage and an answer, generates a question.
\item A \emph{filtering} QA model $p_{f}(a \mid q, C)$ that generates an answer for a given question. If an answer generated by $p_{f}$ does not match the answer a question was generated from, the question is discarded. This ensures generated questions are \emph{consistent}~\cite{alberti_synthetic_2019}.
\end{enumerate}

As shown in~\cref{fig:generation_pipeline}, these models are applied sequentially to generate QA-pairs for \KB{}, in a similar manner to related work in \emph{contextual} QA generation~\cite{alberti_synthetic_2019, lewis_unsupervised_2019}. 
First a passage $c$ is selected with a high probability according to $p_{s}$.
Next, candidate answers $a$ are extracted from $c$ using $p_{a}$, and questions $q$ are generated for each answer in the passage using $p_{q}$. 
Lastly, $p_{f}$ generates a new answer $a'$ for the question. If the source answer $a$ matches $a'$, then $(q,a)$ is deemed consistent and is added to \KB{}.
In the following, we describe each component in detail.
\subsection{Passage Selection, $p_s$} \label{ssec:passage}

The passage selection model $p_s$ is used to find passages which are likely to contain information that humans may ask about, and would thus be good candidates to generate questions from.
We learn $p_s$ using a similar method to \citet{karpukhin_dense_2020}.
Concretely, we assume access to a set of positive passages $C^{+} \subset C$, which we obtain from answer-containing passages from an ODQA training set.
Since we do not have a set of labelled negative passages, we sample negatives from the corpus, either randomly or using heuristics. 
We then train a model to minimise negative log-likelihood of positive passages relative to  negatives.

We implement $p_s$ with RoBERTa~\cite{liu2019roberta} and obtain positive passages from Natural Questions~\cite[NQ,][]{DBLP:journals/tacl/KwiatkowskiPRCP19}. 
We sample \emph{easy negatives} at random from Wikipedia, and \emph{hard negatives} from the same Wikipedia article as the positive passage. 
Easy negatives help the model to learn topics of interest, and hard negatives help to differentiate between interesting and non-interesting passages from the same article.
We evaluate by measuring how highly positive validation passages are ranked amongst negatives.

\subsection{Answer Extraction, $p_a$} \label{ssec:answer}
Given a passage, this component identifies spans that are likely to be answers to questions.
We consider two alternatives: an off-the-shelf Named Entity Recogniser (NER) or training a BERT~\cite{devlin_bert:_2019} answer extraction model on NQ.

The NER answer extractor simply extracts all named entities from a passage.\footnote{We use the sPaCy~\cite{spacy2} NER system, trained on OntoNotes~\cite{hovy_ontonotes_2006}.}
The majority of questions in ODQA datasets consist of entity mentions~\cite{kwiatkowski_natural_2019, joshi_triviaqa:_2017}, so this approach can achieve high answer coverage. 
However, as we extract all entity mentions in a passage, we may extract unsuitable mentions, or miss answers that do not conform to the NER system's annotation schema. 
The trained answer span extractor aims to address these issues.

BERT answer span extraction is typically performed by modelling answer start and end independently, obtaining answer probabilities via $p_a(a \mid c) = p(a_{\text{start}} \mid c) \times p(a_{\text{end}} \mid c)$~\cite{devlin_bert:_2019}. 
We found this approach be sub-optimal for modelling multiple span occurrences in a passage. 
We instead use an approach that breaks the conditional independence of answer spans by directly predicting  $p_a(a \mid c) = p([a_{\text{start}}, a_{\text{end}}] \mid  c)$.
This model first feeds a passage through BERT, before concatenating the start and end token representations of all possible spans of up to length 30, before feeding them into a MLP to compute $p_a(a \mid c)$.
At generation time, the answer extraction component extracts a constant number of spans from each passage, ranked by their extraction probabilities.

\subsection{Question Generation, $p_q$} \label{ssec:question}
Given a passage and an answer, this component generates likely questions with that answer.
To indicate the answer and its occurrence in the passage, we prepend the answer to the passage and label the answer span with surrounding special tokens. 
We construct the dataset from NQ, TriviaQA, and SQuAD, and perform standard fine-tuning of BART-base~\cite{lewis_bart_2019} to obtain $p_q$.

\subsection{Filtering, $p_f$} \label{ssec:filtering}
The filtering model $p_{f}$ improves the quality of generated questions, by ensuring that they are \emph{consistent}: that the answer they were generated is likely to be a valid answer to the question.
Previous work~\cite{alberti_synthetic_2019,fang_accelerating_2020} has employed a machine reading comprehension (MRC) QA model for this purpose, $p_f(a \mid q, c)$, which produces an answer when supplied with a question \emph{and} the passage it was generated from. 
We refer to this as \emph{local filtering}.
However, local filtering will not remove questions which are ambiguous~\cite{min_ambigqa_2020}, and can only be answered correctly with access to the source passage.
Thus, we use an ODQA model for filtering, $p_f(a \mid q, C)$, supplied with only the generated question, and \emph{not} the source passage.
We refer to this as \emph{global filtering}, and later show that it is vital for strong downstream results.
We use FiD-base with 50 retrieved passages, trained on NQ~\cite{izacard_leveraging_2020}.
\section{Question Answering using \KB{}}
\label{sec:qa_with_paq}

We consider two uses of \KB{} for building QA models.
The first is to use \KB{} as a source of training QA-pairs for 
CBQA models. 
The second treats \KB{} as a KB, which models learn to directly retrieve from.
These use-cases are related, as CBQA models have been shown to memorise the train data in their parameters, latently retrieving from them at test time~\cite{lewis_question_2020,domingos_every_2020}.

\subsection{\KB{} for Closed-Book QA}

We fine-tune a BART-large~\cite{lewis_bart_2019} with QA-pairs from the concatenation of the training data and \KB{}, using a similar training procedure to \citet{roberts_how_2020}.
We use early stopping on the validation set and a batch size of 512, and note learning is slow, requiring 70 epochs on \KB{}.
Following recent best practices\cite{alberti_synthetic_2019,yang_data_2019}, we then fine-tune on the training QA-pairs only, using validation Exact Match score for early stopping~\cite{rajpurkar_know_2018}.

We note that an effective CBQA model must be able to understand the semantics of questions and how to generate answers, in addition to being able to store a large number of facts in its parameters.
This model thus represents a kind of combined  \emph{parametric} knowledgebase and retrieval system~\cite{petroni_how_2020}.
The model proposed in the next section, \QAEngine{}, represents an explicit \emph{non-parametric} 
instantiation of this idea.

\subsection{\QAEngine{}}
\QAEngine{} is a retrieval model which operates on KBs of QA-pairs, such as \KB{}. 
\QAEngine{} extends recently proposed nearest neighbour QA-pair retriever models~\cite{lewis_question_2020,xiao_open-domain_2020}.
These models assume access to a KB of $N$ QA-pairs  $\mathcal{K} = \{(q_{1},a_{1}) ... (q_{N},a_{N})\}$. 
These models provide an answer to a test question $q$ by finding the most relevant QA-pair $(q',a') $ in $\mathcal{K}$, using a scalable  relevance function, then returning $a'$ as the answer to $q$. 
Such a function could be implemented using standard information retrieval techniques, (e.g. TF-IDF) or learnt from training data.
\QAEngine{} is learnt from ODQA data and consists of a neural dense retriever, optionally followed by a neural reranker. 

\subsubsection{\QAEngine{} Retriever}
Our retriever adopts the dense Maximum Inner Product Search (MIPS) retriever paradigm, that has recently been shown to obtain state-of-the-art results in a number of settings~\cite[inter alia]{karpukhin_dense_2020, lee_learning_2021}. 
Our goal is to embed queries $q$ and indexed items $d$ into a representation space 
via embedding functions $g_{q}$ and $g_{d}$, so that the inner product $g_{q}(q)^{\top} g_{d}(d)$ is maximised for items relevant to $q$.
In our case, queries are questions and indexed items are QA-pairs $(q',a')$. 
We make our retriever symmetric by embedding  $q'$ rather than $(q',a')$, meaning that \emph{only one} embedding function $g_q$ is required, which maps questions to embeddings. This applies a useful inductive bias, and we find that it aids stability during training.

Learning the embedding function $g_q$ is complicated by the lack of labelled question pair paraphrases in ODQA datasets. 
We propose a latent variable approach similar to retrieval-augmented generation~\cite[RAG,][]{lewis_question_2020},\footnote{Other methods, such as heuristically constructing paraphrase pairs assuming that questions with the same answer are paraphrases, and training with sampled negatives would also be valid, but were not competitive in our early experiments} where we we index training QA-pairs rather than documents.
For an input question $q$, the top $K$ QA-pairs $(q',a')$ are retrieved by a retriever $p_{\texttt{ret}}$ where $p_{\texttt{ret}}(q | q') \propto \exp(g_q(q)^{\top{}} g_q(q'))$.
These are then fed into a seq2seq model $p_{\texttt{gen}}$ which generates an answer for each retrieved QA-pair, before a final answer is produced by marginalising,
\[
p(a|q)=\sum_{\mathclap{(a',q') \in \text{top-}k  \ p_{\texttt{ret}}(\cdot|q)}} p_{\texttt{gen}}(a| q, q', a')p_{\texttt{ret}}(q'|q), 
\]
As $p_{\texttt{gen}}$ generates answers token-by-token, credit can be given for retrieving helpful QA-pairs which do not exactly match the target answer. 
For example, for the question ``when was the last time anyone was on the moon'' and target answer ``December 1972'', retrieving ``when was the last year astronauts landed on the moon'' with answer ``1972'' will help to generate the target answer, despite the answers having different granularity.
After training, we discard $p_{\texttt{ret}}$\footnote{We could use $p_{\texttt{gen}}$ as a reranker/aggregator for QA, but in practice find it both slower and less accurate than the reranker described in \cref{sec:reranker}}, retaining only the question embedder $g$.
We implement $p_{\texttt{ret}}$ with ALBERT~\cite{Lan2020ALBERT} with an output dimension of 768, and $p_{\texttt{gen}}$ with BART-large~\cite{lewis_bart_2019}. 
We train using the top 100 retrieved QA-pairs, and refresh the embedding index every 5 training steps.

Once the embedder $g_q$ is trained, we build a test-time QA system by embedding and indexing a QA KB such as \KB{}.
Answering is achieved by retrieving the most similar stored question, and returning its answer. 
The matched QA-pair can be displayed to the user, providing a mechanism for more interpretable answers than CBQA models and many retrieve-and-read generators which consume thousands of tokens to generate an answer.
Efficient MIPS libraries such as FAISS~\cite{johnson_billion-scale_2017} enable \QAEngine{}'s retriever to answer 100s to 1000s of questions per second (see \cref{sec:speed}).
We use a KB for \QAEngine{} consisting of training set QA-pairs concatenated with QA-pairs from \KB{}.

\subsubsection{\QAEngine{} Reranker}
\label{sec:reranker}

The accuracy of \QAEngine{} can be improved using a reranker on the top-$K$ QA-pairs from the retriever.
The reranker uses
cross-encoding~\cite{humeau_poly-encoders_2020}, and includes the retrieved answer in the scoring function for richer featurisation.
We concatenate the input question $q$, the retrieved question $q'$ and its answer $a'$ with separator tokens, and feed it through ALBERT. 
We obtain training data in the following manner: For a training QA-pair, we first retrieve the top $2K$ QA-pairs from \KB{} using \QAEngine{}'s retriever. 
If one of the retrieved QA-pairs has the correct answer, we treat that QA-pair as a positive, and randomly sample K-1 of the incorrect retrieved questions as negatives. 
We train by minimising negative log likelihood of positives relative to 10 negatives, and rerank 50 retrieved pairs at test time. 
The reranker improves accuracy at the expense of some speed. However, as QA-pairs consist of  fewer tokens than passages, the reranker is still faster than retrieve-and-read models, even when using architectures such as ALBERT-xxlarge. 
\section{Results}

We first examine the \KB{} resource in general, before exploring how both CBQA models and \QAEngine{} perform using \KB{}, comparing to recently published systems. 
We use the Natural Questions~\citep[NQ,][]{DBLP:journals/tacl/KwiatkowskiPRCP19} and TriviaQA~\cite{joshi_triviaqa:_2017} datasets to assess performance, evaluating with the standard Exact Match (EM) score.

\subsection{Examining PAQ}

We generate \KB{} by applying the  pipeline described in \cref{sec:generating_question_answer_pairs}
to the Wikipedia dump from \citet{DBLP:conf/emnlp/KarpukhinOMLWEC20}, which splits Wikipedia into 100-word passages.
We use passage selection model $p_s$ to rank all 21M passages, and generate from the top 10M, before applying global filtering.\footnote{Generation was stopped when downstream performance with \QAEngine{} did not significantly improve with more questions.} 

We are interested in understanding the effectiveness of different answer extractors,
and whether generating more questions per answer span results leads to better results.
To address these questions, we create three versions of PAQ, described below.
\KB{}$_{L}$ uses the learnt answer extractor, and a question generator trained on NQ and TriviaQA. We extract 8 answers per passage and use a beam size of 4 for question generation.
In \KB{}$_{L,1}$ we only use the top scoring question in the beam, whereas in \KB{}$_{L,4}$ we use all four questions from the beam,  allowing for several questions to be generated from one answer in a passage. 
\KB{}$_{NE,1}$ uses the NER answer extractor, and a generator trained only on NQ.
\KB{}$_{NE,1}$ allow us assess whether diversity in the form of answer extractors and question generators leads to better results.
The final KB, referred to as just ``\KB{}'', is the union of \KB{}$_L$ and \KB$_{NE}$. 
 
As shown in \cref{tab:paq_stats},  \KB{} consists of 65M filtered QA pairs.\footnote{
Each question only has one answer due to global filtering
} This was obtained by extracting 165M answer spans and generating 279M unique questions before applying global filtering.
\cref{tab:paq_stats} shows that the \KB{}$_L$ pipeline is more efficient than \KB{}$_{NE}$, with 24.4\% of QA-pairs surviving filtering, compared to 18\%. 

\begin{table}[t]
\centering
\setlength{\tabcolsep}{3pt}
\small
\begin{tabular}{@{}lcccc|cc@{}}
\toprule[1pt]

\multirow{2}{0.1cm}{Dataset} &\multirow{2}{1.1cm}{\centering Extracted Answers} & \multirow{2}{1.0cm}{\centering Unique Qs} &  \multirow{2}{0.9cm}{\centering Filtered QAs}&  \multirow{2}{0.8cm}{Ratio}&\multicolumn{2}{c}{Coverage} \\
 & & & & & \multicolumn{1}{c}{NQ} & \multicolumn{1}{c}{TQA} \\
\midrule
\KB{}$_{L,1}$ & 76.4M  & 58.0M  & 14.1M & 24.4\%  & 88.3&	90.2 \\
\KB{}$_{L,4}$  & 76.4M  & 225.2M & 53.8M & 23.9\% & 89.8&	90.9 \\
\KB{}$_{NE,1}$ & 122.2M & 65.4M  & 12.0M & 18.6\% & 83.5&	88.3 \\ \midrule 
PAQ           & 165.7M & 279.2M   &  64.9M & 23.2\%    & 90.2&	91.1 \\
\bottomrule
\end{tabular}
\caption{PAQ dataset statistics and ODQA dataset answer coverage. ``Ratio'' refers to the number of generated questions which pass the global consistency filter.}
\label{tab:paq_stats}

\end{table}
\begin{table*}[ht]
\centering
\hskip-2mm
\scalebox{0.85}{
\setlength{\tabcolsep}{1pt}
\begin{tabular}{lllll}
\toprule[1pt]
\# & Question  & Answer & Comment \\
\midrule
1 &who created the dutch comic strip panda & Martin Toonder &\checkmark \\
2 &what was the jazz group formed by john hammond in 1935 & Goodman Trio & \checkmark\\
3 &astrakhan is russia's main market for what commodity & fish & \checkmark\\
4 &what material were aramaic documents rendered on & leather & \checkmark  \\
5 &when did the giant panda chi chi died & 22 July 1972 & \checkmark, Grammar error\\
6 &pinewood is a village in which country & England & $\sim$,\ Also a Pinewood village in USA  \\
7 &who was the mughal emperor at the battle of lahore & Ahmad Shah Bahadur & \xmark\  Confuses with Ahmad Shah Abdali \\
8 & how many jersey does mitch richmond have in the nba & 2 & \xmark\  His Jersey No. was 2 \\
\bottomrule
\end{tabular}}
\caption{Representative Examples from PAQ. \checkmark indicates correct, $\sim$ ambiguous and \xmark{} incorrect facts respectively}
\label{tab:paq_examples}
\end{table*}
\paragraph{PAQ Answer Coverage}
To evaluate answer extractors, we calculate how many answers in the validation sets of TriviaQA and NQ also occur in \KB{}'s filtered QA-pairs.\footnote{performed using standard answer normalisation~\cite{rajpurkar_squad:_2016}}
Table \ref{tab:paq_stats} shows that the answer coverage of \KB{} is very high -- over 90\% for both TriviaQA and NQ. %
Comparing \KB{}$_{L}$ with \KB{}$_{NE}$ shows that the learnt extractor achieves higher coverage, but the union of the two leads to the highest coverage overall. Comparing \KB{}$_{L,1}$ and  \KB{}$_{L,4}$ indicates that using more questions from the beam also results in higher coverage. 

\paragraph{\KB{} Question Generation Quality}

Illustrative examples from \KB{} can be seen in \cref{tab:paq_examples}. 
Manual inspection of 50 questions from \KB{} reveals that 82\% of questions accurately capture information from the passage and contain sufficient details to locate the answer. 
16\% of questions confuse the semantics of certain answer types, either by conflating  similar entities in the passage or by misinterpreting rare phrases (see examples 7 and 8 in \cref{tab:paq_examples}).
Finally, we find small numbers of grammar errors (such as example 5) and mismatched wh-words (5\% and 2\% respectively).\footnote{Further details and automatic metrics in \cref{appendix:qgen_assessment}}
\paragraph{Other observations}
PAQ often contains several paraphrases of the same QA-pair.
This redundancy reflects how information is distributed in Wikipedia, with facts often mentioned on several different pages. 
Generating several questions per answer span also increases redundancy.  
Whilst this means that \KB{} could be more information-dense if a de-duplication step was applied, we later show that \QAEngine{} always improves with more questions in its KB (section $\ref{sec:ablating_paq_with_repaq}$).
This suggests that it is worth increasing redundancy for greater coverage.

\subsection{Question Answering Results}

\begin{table*}[t]
\centering
\setlength{\tabcolsep}{5pt}
\scalebox{0.88}{
\begin{tabular}{@{}lllcc@{}}
\toprule[1pt]
\#&Model Type & 
\multirow{1}{*}{Model} 
& \multicolumn{1}{c}{NaturalQuestions}
& \multicolumn{1}{c}{TriviaQA} \\
\midrule
1&Closed-book & T5-11B-SSM~\cite{roberts_how_2020} & 35.2 & 51.8\\
2&Closed-book & BART-large~\cite{lewis_question_2020} & 26.5 & 26.7\\
3&QA-pair retriever & Dense retriever~\cite{lewis_question_2020} & 26.7 & 28.9\\
4&Open-book, retrieve-and-read & RAG-Sequence~\cite{lewis_retrieval-augmented_2020} & 44.5 & 56.8\\
5&Open-book, retrieve-and-read & FiD-large, 100 docs~\cite{izacard_leveraging_2020} & 51.4 & \textbf{67.6}\\
6&Open-book, phrase index & DensePhrases~\cite{lee_learning_2021}& 40.9  & 50.7 \\
\midrule 
7&Closed-book & BART-large, pre-finetuned on \KB{} & 32.7  & 33.2 \\
8&QA-pair retriever & \QAEngine{} (retriever only) & 41.2 & 38.8 \\
9&QA-pair retriever & \QAEngine{} (with reranker)  & \underline{47.7} & 50.7 \\
10&QA-pair retriever & \QAEngine{}-multitask (retriever only) & 41.7 & 41.3\\
11&QA-pair retriever & \QAEngine{}-multitask (with reranker) & 47.6 & \underline{52.1}\\
12&QA-pair retriever & \QAEngine{}-multitask w/ FiD-Large Backoff & \textbf{52.3} & 67.3\\

\bottomrule
\end{tabular}
}
\caption{Exact Match score for highest accuracy \QAEngine{} configurations in comparison to recent state-of-the-art systems. 
Highest score indicated in bold, highest  non-retrieve-and-read model underlined.
}
\label{tab:main_results}
\end{table*}

In this section, we shall compare how the \KB{}-leveraging models proposed in section \ref{sec:qa_with_paq} compare to existing approaches.
We primarily compare to a state-of-the-art retrieve-and-read model, Fusion-in-Decoder \cite[FiD,][]{izacard_leveraging_2020}.
FiD uses DPR~\cite{karpukhin_dense_2020} to retrieve relevant passages from Wikipedia, and feeds them into T5~\cite{raffel_exploring_2020} to generate a final answer. 

Table \ref{tab:main_results} shows the highest-accuracy configurations of our models alongside recent state-of-the-art approaches.
We make the following observations:
Comparing rows 2 and 7 shows that a CBQA BART model trained with \KB{} outperforms a comparable NQ-only model by 5\%, and only 3\% behind T5-11B (row 1) which has 27x more parameters.
Second, we note strong results for \QAEngine{} on NQ (47.7\%, row 9), outperforming recent retrieve-and-read systems such as RAG by over 3\% (row 4).

Multi-task training \QAEngine{} on NQ and TriviaQA improves TriviaQA results by 1-2\% (comparing rows 8-9 with 10-11).  
\QAEngine{} does not perform quite as strongly on TriviaQA (see section \ref{sec:does_filtering_limit_qa}), 
but is within 5\% of RAG, and outperforms concurrent work on real-time QA, DensePhrases~\cite[row 6,][]{lee_learning_2021}. 
Lastly, row 12 shows that combining \QAEngine{} and FiD-large into a combined system is 0.9\% more accurate than FiD-large (see Section \ref{sec:know_when_you_dont_know} for more details).

\subsubsection{Ablating \KB{} using \QAEngine{}}
\label{sec:ablating_paq_with_repaq}
\cref{tab:PAQ_ablations} shows \QAEngine{}'s accuracy when using different \KB{} variants. 
To establish the effect of filtering, we evaluate \QAEngine{} with unfiltered, locally-filtered and globally-filtered QA-pairs on \KB{}$_{L,1}$.
Comparing rows 2, 3 and 4 shows that global filtering is crucial, leading to a 9\% and 14\% improvement over locally-filtered and unfiltered datasets respectively.

We also note a general trend in Table \ref{tab:PAQ_ablations} that adding more globally-filtered questions improves accuracy.
Rows 4 and 5 show that using four questions per answer span is better than generating one (+0.9\%), and 
Rows 5,6 and 7 show that combining \KB$_{NE}$ and \KB$_{L}$ results in a further 1.2\% improvement.
Empirically we did not observe any cases where increasing the number of globally filtered QA-pairs reduced accuracy, even when there were millions of QA-pairs already.

\begin{table}[t]
\centering
\setlength{\tabcolsep}{5pt}
\scalebox{0.9}{
\begin{tabular}{@{}lllccc@{}}
\toprule[1pt]
\multirow{2}{*}{\#}&\multirow{2}{*}{KB} & \multirow{2}{*}{Filtering} & \multirow{2}{*}{Size} & \multicolumn{2}{c}{Exact Match} \\
& & & & \multicolumn{1}{c}{Retrieve} & \multicolumn{1}{c}{Rerank} \\

\midrule
1 & NQ-Train & \multicolumn{1}{c}{-} & 87.9K & 27.9 & 31.8\\
\midrule
2 & \KB{}$_{L,1}$ & None & 58.0M & 21.6 & 30.6 \\
3 & \KB{}$_{L,1}$ & Local & 31.7M & 28.3 & 34.9\\
4 & \KB{}$_{L,1}$ & Global & 14.1M& 38.6 & 44.3\\
5 & \KB{}$_{L,4}$ & Global & 53.8M & 40.3 & 45.2\\
6 & \KB{}$_{NE,1}$ & Global & 12.0M& 37.3 & 42.6 \\
\midrule 
7 & \KB{} & Global & 64.9M & \textbf{41.6} & \textbf{46.4}\\
\bottomrule
\end{tabular}
}
\caption{The effect of different \KB{} subsets on the NQ validation accuracy of \QAEngine{}}
\label{tab:PAQ_ablations}
\end{table}

\subsubsection{System Size vs Accuracy}
\label{sec:system_size_v_accuracy}

\KB{}'s QA-pairs are accompanied by scores of how likely they are to be asked.
These scores can be used to filter the KB and reduce the \QAEngine{} system size.
A similar procedure can be used to filter the background corpus for a retrieve-and-read model~\cite{izacard_memory_2020}.
We compare the system size of a FiD-large system and \QAEngine{} as the number of items (passages and QA-pairs respectively) in their indexes are reduced. 
We select which passages and QA-pairs are included using the passage selection model $p_s$.\footnote{Here, we use \KB{}$_{L1}$, which is 5x smaller than the full \KB{}, but retains most of the accuracy (see Table~\ref{tab:PAQ_ablations})}
Further experimental details can be found in appendix \ref{appendix:memory}.
\cref{fig:memory_plot} shows the that both system sizes can be reduced several-fold with only a small drop in accuracy, demonstrating the effectiveness of $p_s$.
FiD can achieve a higher accuracy, but requires larger system sizes.
\QAEngine{} can be reduced to a smaller size before a significant accuracy drop,  driven primarily by the higher information density of QA-pairs relative to passages, and fewer model parameters used by \QAEngine{} compared to FiD.
Highly-optimized \QAEngine{} models won the ``500MB'' and ``Smallest System" tracks of the EfficientQA NeurIPS competition with disk images of 336MB and 29MB respectively.
For further details on EfficientQA, the reader is referred to \citet{min_neurips_2020}.

\begin{figure}[t]
\centering
  \includegraphics[width=0.48\textwidth]{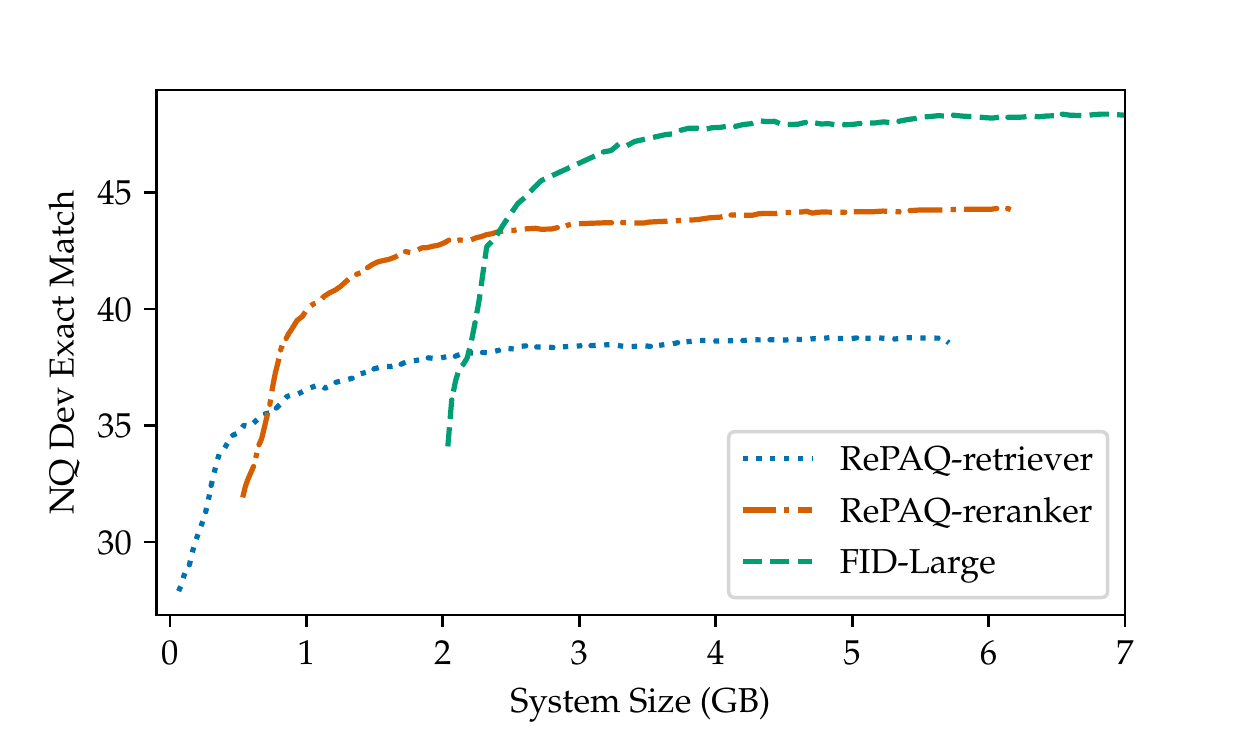}
  \caption{System size vs. accuracy for \QAEngine{}  and FiD-large as a function of the number of items in the index.}
  \label{fig:memory_plot}
\end{figure}

\subsubsection{Inference Speed vs Accuracy} 
\label{sec:speed}

We train a variety of differently-sized \QAEngine{} models to explore the relationship between accuracy and inference speed.
We use a fast Hierarchical Navigable Small World~(HNSW) index in FAISS \cite{malkov_efficient_2018, johnson_billion-scale_2017}\footnote{The HNSW index has negligible ($\sim$0.1\%) drop in retriever accuracy compared to a flat index} and measure the time required to evaluate the NQ test set on a system with access to one GPU.\footnote{System details can be found in Appendix \ref{appendix:speed}}
Table \ref{tab:speed} shows these results.
Some retriever-only \QAEngine{} models can answer over 1000 questions per second, and are relatively insensitive to model size, with ALBERT-base only scoring 0.5\% lower than ALBERT-xlarge. 
They also outperform retrieve-and-read models like REALM~\cite[40.4\%,][]{guu_realm_2020} and recent  real-time QA models like DensePhrases~\cite[40.9\%,][]{lee_learning_2021}. 
We find that larger, slower \QAEngine{} rerankers achieve higher accuracy. However, even the slowest \QAEngine{} is 3x faster than FiD-base, whilst only being 0.8\% less accurate, and  12x faster than FiD-large.

\begin{table}[t]
\centering
\setlength{\tabcolsep}{5pt}
\scalebox{0.85}{
\begin{tabular}{@{}lcccc@{}}
\toprule[1pt]
Model & Retriever & Reranker & Exact Match & Q/sec \\
\midrule
\multicolumn{1}{l}{FiD-large} & - & - & 51.4 & 0.5\\
\multicolumn{1}{l}{FiD-base} & - & - & 48.2 & 2\\ \midrule
\multicolumn{1}{l}{\QAEngine{}} &  base & - & 40.9 & 1400 \\
\multicolumn{1}{l}{\QAEngine{}} &  large & - & 41.2 & 1100\\
\multicolumn{1}{l}{\QAEngine{}} &  xlarge & - & 41.5  & 800 \\ \midrule 
\multicolumn{1}{l}{\QAEngine{}} &  base & base &  45.7 & 55\\
\multicolumn{1}{l}{\QAEngine{}} &  large & xlarge & 46.2 &  10 \\
\multicolumn{1}{l}{\QAEngine{}} &  xlarge & xxlarge & 47.6 &  6 \\
\bottomrule
\end{tabular}
}
\caption{Inference speeds of various configurations of \QAEngine{} compared to FiD models on NQ
} 
\label{tab:speed}
\end{table}

\subsubsection{Selective Question Answering} 
\label{sec:know_when_you_dont_know}
\begin{figure}[t]
\centering
  \includegraphics[width=0.48\textwidth]{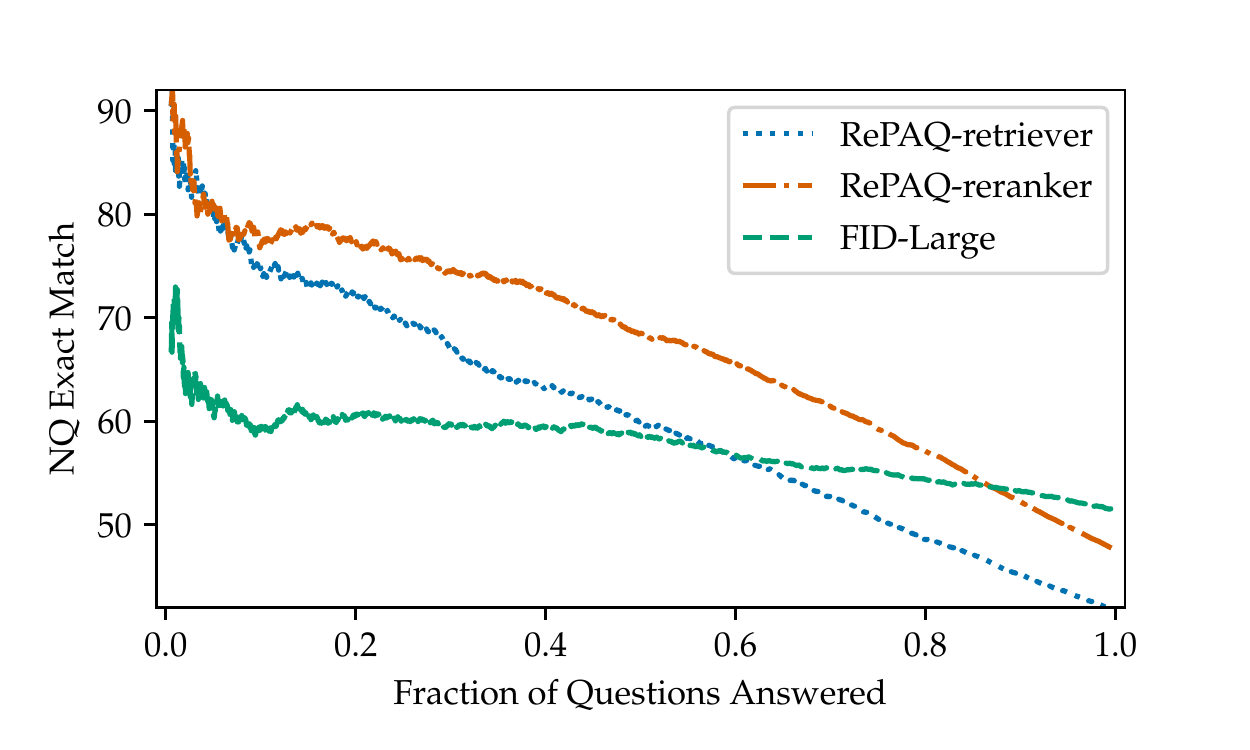}
  \caption{Risk-coverage plot for FiD and \QAEngine{}.}
  \label{fig:risk_plot}
\end{figure}
QA systems should not just be able to answer accurately, but also ``know when they don't know", and abstain from answering when they are unlikely to produce good answers.
This task is challenging for current systems~\cite{asai_challenges_2020,jiang_generalizing_2020}, and has been approached in MRC by training on unanswerable questions~\cite{rajpurkar_know_2018} and for trivia systems by leveraging incremental QA formats~\cite{rodriguez_quizbowl_2019}.

We find that \QAEngine{}'s retrieval and reranking scores are well-correlated with answering correctly. 
This allows \QAEngine{} to be used for selective question answering by abstaining  when the score is below a certain threshold. 
Figure \ref{fig:risk_plot} shows a \emph{risk-coverage} plot~\cite{wang_safer_2018} for \QAEngine{} and FiD, where we use FiD's answer log probability for its answer confidence.\footnote{We also investigate improving FiD's calibration using an auxiliary model, see Appendix \ref{appendix:selective_qa}. We find that the most effective way to calibrate FiD is to use \QAEngine{}'s confidences} 
The plot shows the accuracy on the top N\% highest confidence answers for NQ. 
If we require models to answer 75\% of user questions, \QAEngine{}'s accuracy on the questions it does answer is 59\%, whereas FiD, which has poorer calibration, scores only 55\%. 
This difference is even more pronounced with stricter thresholds -- with coverage of 50\%, \QAEngine{} outperforms FiD by over 10\%.
FiD only outperforms \QAEngine{} when we require systems to answer more than 85\% of questions.

Whilst \QAEngine{}'s selective QA is useful in its own right, it also allows us to combine the slow but accurate FiD with the fast and precise \QAEngine{}, which we refer to as \emph{backoff}.
We first try to answer with \QAEngine{}, and if the confidence is below a threshold determined on validation data, we pass the question onto FiD.
For NQ, the combined system is 2.1x faster than FiD-large, with \QAEngine{} answering 57\% of the questions, and the overall accuracy is 1\% higher than FiD-large (see table \ref{tab:main_results}). 

If inference speed is a priority, the threshold can be decreased so that \QAEngine{} answers 80\% of the questions, which retains the same overall accuracy as FiD, with a 4.6x speedup. 
For TriviaQA, the combined system backs off to FiD earlier, due to the stronger relative performance of FiD.
Additional details can be found in appendix \ref{appendix:selective_qa}

\subsubsection{Analysing \QAEngine{}'s Predictions} 
\label{sec:error_analysis}
Some examples of top retrieved questions are shown \cref{tab:repaq_examples}.
When \QAEngine{} answers correctly, the retrieved question is a paraphrase of the test question from \KB{} in 89\% of cases.  
As such, there is high (80.8 ROUGE-L) similarity between correctly-answered test questions and the top retrieved questions.
9\% of test questions even exist verbatim in \KB{}, and are thus trivial to answer.
The reranker primarily improves over the retriever for ambiguous cases, and cases where the top retrieved answer does not have the right granularity.
In 32\% of cases, \QAEngine{} does not retrieve the correct answer in the top 50 QA-pairs, suggesting a lack of coverage may be a significant source of error. In these cases, retrieved questions are much less similar to the test question than for correctly answered questions, dropping by 20 ROUGE-L. 
We also observe cases where retrieved questions match the test question, but the retrieved answer does not match the desired answer. This is usually due to different answer granularity, but in a small number of cases was due to factually incorrect answers.
\subsubsection{Does the Filtering Model Limit \QAEngine{}'s Accuracy?}
\label{sec:does_filtering_limit_qa}

As \QAEngine{} relies on retrieving paraphrases of test questions, we may expect that the ODQA filtering model places an upper bound on it's performance.
For example, if a valid QA-pair is generated which overlaps with a test QA-pair, but the filter cannot answer it correctly, that QA-pair will not be added to \KB{}, and \QAEngine{} cannot use it to answer the test question.
The NQ FiD-base-50-passage model used for filtering scores 46.1\% and 53.1\% for NQ and TriviaQA respectively.
\QAEngine{} actually outperforms the filter model on NQ by 1.6\%.
This is possible because generated questions may be phrased in such a way that they are easier to answer, e.g. being less ambiguous~\cite{min_ambigqa_2020}. 
\QAEngine{} \emph{can} then retrieve the paraphrased QA-pair and answer correctly, even if the filter could not answer the test question directly.
The filtering model's weaker scores on TriviaQA helps explain why \QAEngine{} is not as strong on this dataset.
We speculate that using a stronger filtering model for TriviaQA would in turn improve \QAEngine{}'s results. 

\begin{table}[t]
\hskip-2mm
\scalebox{0.7}{

\setlength{\tabcolsep}{1pt}
\begin{tabular}{lll}
\toprule[1pt]
\textbf{Input}: who was the film chariots of fire about                           &  \textbf{A}: Eric Liddell\\
\midrule
 \emph{who was the main character in chariots of fire}                 &  \textbf{A}: \emph{Eric Liddell} & \checkmark{}\\
 who starred in the movie chariots of fire                             &  \textbf{A}: {\fontsize{9}{11}\selectfont Ian Charleson}       & \xmark\\
 which part did straan rodger play in chariots of fire            &  \textbf{A}: {\fontsize{9}{11}\selectfont Sandy McGrath}& \xmark\\
 who played harold in the 1981 film chariots of fire          &  \textbf{A}: Ben Cross           & \xmark{}\\
 who is the main character in chariots of fire                         &  \textbf{A}: Eric Liddell        & \checkmark{} \\
\midrule
\textbf{Input}: what is the meaning of the name didymus      &  \textbf{A}: twin\\
\midrule
 what language does the name didymus come from    &  \textbf{A}: Greek         & \xmark{} \\
 where does the name didymus come from in english &  \textbf{A}:  Greek        & \xmark{} \\ 
 what does the word domus mean in english         &  \textbf{A}:  home         & \xmark{} \\
 how long has the term domus been used            &  \textbf{A}: {\fontsize{9}{11}\selectfont 1000s of years}& \xmark{}\\
 \emph{what does the greek word didyma mean}      &  \textbf{A}: \emph{twin}   & \checkmark{}\\
\midrule
 \textbf{Input}: what is the name of a group of llamas &   \textbf{A}: herd \\
\midrule
 what are llamas and alpacas considered to be &   \textbf{A}: {\fontsize{9}{11}\selectfont domesticated} & \xmark{}\\
 what are the figures of llamas in azapa valley &   \textbf{A}: Atoca & \xmark{}\\
 what are the names of the llamas in azapa valley &  \textbf{A}: Atoca & \xmark{}\\
 \emph{what is the scientific name for camels and llamas} &   \textbf{A}:\emph{Camelidae} & \xmark{}\\
 are llamas bigger or smaller than current forms &   \textbf{A}:larger & \xmark{}\\
\bottomrule
\end{tabular}
}
\caption{Examples of top 5 retrieved QA-pairs for NQ. Italics indicate QA-pairs chosen by reranker.}
\label{tab:repaq_examples}
\end{table}

\begin{table}[t]
\centering
\hskip-2mm
\scalebox{0.78}{
\setlength{\tabcolsep}{0pt}
\begin{tabular}{lcccc}
\toprule[1pt]
Model & \multicolumn{1}{m{1cm}}{\centering Total} &\multicolumn{1}{m{1.5cm}}{\centering Q Overlap} & \multicolumn{1}{m{1.5cm}}{\centering A-only Overlap}&  \multicolumn{1}{m{1.5cm}}{\centering No Overlap}\\ \midrule
CBQA BART w/ NQ & 26.5 & 67.6 &  10.2& 0.8 \\\midrule
CBQA BART w/ NQ+PAQ& 28.2 & 52.8 & 24.4 & 9.4\\
\ \ \ \ \ \ + final NQ finetune  & 32.7 & 69.8 & 22.2 & 7.51\\
\midrule
\QAEngine{} (retriever only) & 41.7 & 65.4 & 31.7 & 21.4\\
\QAEngine{} (with reranker) & 47.3& 73.5&39.7 &26.0\\
\bottomrule
\end{tabular}
}
\caption{NQ Behavioural splits~\cite{lewis_question_2020}. ``Q overlap'' are test questions with  paraphrases in training data. ``A-only'' are test questions where answers appear in training data, but questions do not. ``No overlap'' where neither question or answer overlap.
} 
\label{tab:overlap}
\end{table}

\subsection{Closed-book QA vs \QAEngine{}} 
Table \ref{tab:overlap} shows results on test set splits which measure how effectively models memorise QA-pairs from the NQ train set (``Q overlap''), and generalise to novel questions (``A overlap only'' and ``No overlap'').\footnote{See \citet{lewis_question_2020} for further details.}
Comparing CBQA models trained on NQ vs those trained on NQ and \KB{} show that models trained with \KB{} answer more questions correctly from the ``A-only overlap'' and ``No overlap'' categories, indicating they have learnt facts not present in the NQ train set.
Applying additional NQ finetuning on the \KB{} CBQA model improves scores on ``Q overlap'' (indicating greater memorisation of NQ), but scores on the other categories drop (indicating reduced memorisation of \KB{}).
However, \QAEngine{}, which explicitly retrieves from \KB{} rather than memorising it in parameters, strongly outperforms the CBQA model in all categories, demonstrating that the CBQA model struggles to memorise enough facts from \KB{}. 
CBQA models with more parameters may be better able to memorise \KB{}, but have downsides in terms of system resources.
Future work should address how to better store \KB{} in CBQA model parameters.

\section{Related Work}
ODQA has received much attention in both for its practical applications, and as a benchmark for how NLP models store and access knowledge~\cite{DBLP:conf/acl/ChenY20,petroni_kilt_2020}.

\paragraph{KBQA} A number of early approaches in ODQA focused on using structured KBs~\citep{DBLP:conf/emnlp/BerantCFL13} such as Freebase~\citep{DBLP:conf/aaai/BollackerCT07}, with recent examples from \citet{fevry_entities_2020} and \citet{verga_facts_2020-1}. 
This approach often has high precision but suffers when KB doesn't match user requirements, or where the schema limits what knowledge can be stored.
We populate our knowledgebase with semi-structured QA pairs which are specifically likely to be relevant at test time mitigating both of these drawbacks, and sharing many of the benefits, such as precision and extensibility. 

\paragraph{Open Information Extraction} Our work touches on automatic KB construction and open information extraction (OpenIE)~\cite{angeli_leveraging_2015}. 
Here, the goal is to mine facts from free text into structured or semi-structured forms, typically (subject, relation, object) triples for use in tasks such as slot-filling~\cite{surdeanu_overview_2013}.
We generate natural language QA-pairs rather than OpenIE triples, and we do not attempt to extract all possible facts in a corpus, focusing only those that are likely to be asked. QA-pairs have also been used in semantic role labelling, such as QA-SRL~\cite{fitzgerald_large-scale_2018}.

\paragraph{Real-time ODQA} Systems that prioritise fast runtimes over accuracy are sometimes referred to as \emph{real-time QA} systems~\cite{seo_phrase-indexed_2018}. 
DenSPI~\cite{seo_real-time_2019} and a contemporary work, DensePhrases~\cite{lee_learning_2021}, approach this by indexing all possible phrases in a background corpus, and learn mappings from questions to passage-phrase pairs. 
We also build an index for fast answering, but generate and index globally-answerable questions.
Indexing QA-pairs can be considered as indexing summaries of important facts from the corpus, rather than indexing the corpus itself.
We also generate and store multiple questions per passage-answer pair, relieving information bottlenecks from encoding a passage-answer pair into a single vector. 

\paragraph{Question Generation for QA} Question generation has been used for various purposes, such as data augmentation~\cite{alberti_synthetic_2019, lewis_unsupervised_2019,lee_learning_2021}, improved retrieval~\cite{nogueira_document_2019}, generative modelling for contextual QA~\cite{lewis_generative_2018}, as well as being studied in its own right~\cite{du_learning_2017, hosking_evaluating_2019}.
\citet{serban_generating_2016} generate large numbers of questions from Freebase, but do not address how to use them for QA.
Closest to our work is the recently-proposed OceanQA~\cite{fang_accelerating_2020}. OceanQA first generates contextual QA-pairs from Wikipedia.
At test-time, a document retrieval system is used to retrieve the most relevant passage for a question and the closest pre-generated QA-pair from that passage is then selected. 
In contrast, we focusing on generating a large KB of non-contextual, globally-consistent ODQA questions and exploring what QA systems are facilitated by such a resource.

\section{Conclusion}

We have introduced \KB{}, a dataset of 65M QA-pairs, and explored how it could be used to improve ODQA models. 
We demonstrated the effectiveness of \QAEngine{}, a system which retrieves from \KB{}, in terms of accuracy, speed, space efficiency and selective QA.
Generating \KB{} is computationally intensive due to its large scale, but should be a useful, re-usable resource for more accurate, smaller and faster QA models.
Nevertheless, future work should be carried out to improve the efficiency of generation in order to expand \KB{}'s coverage.

We also demonstrated \KB{}'s utility for improved CBQA models, but note a large accuracy gap between our CBQA models and \QAEngine{}. 
Exploring the trade-offs between storing and retrieving knowledge parametrically or non-parametrically is a topic of great current interest~\cite{lewis_retrieval-augmented_2020,de_cao_autoregressive_2020}, and \KB{} should be a useful testbed for probing this relationship further.
We also note that \KB{} could be used as general data-augmentation when training any open-domain QA model or retriever.
Whilst we consider such work out-of-scope for this paper, leveraging \KB{} to improve retrieve-and-read and other systems systems should be explored in future work.

\bibliography{bibliography,patrick_refs}
\bibliographystyle{acl_natbib}
\clearpage
\appendix
\section{Appendices}

\subsection{Dataset splits}
\label{appendix:splits}

For Natural Questions~\citep[NQ,][]{DBLP:journals/tacl/KwiatkowskiPRCP19}, we use the standard Open-Domain splits in our experiments, consisting of 79,168 train, 8,757 development, and 3,610 test question-answer pairs.

For TriviaQA, We use the open-domain train-test splits, which correspond to the unfiltered-train and unfiltered-dev reading comprehension splits~\citep{DBLP:conf/acl/LeeCT19,DBLP:conf/emnlp/MinCHZ19,DBLP:conf/emnlp/KarpukhinOMLWEC20}. 

\subsection{Futher details on Passage selection}
\label{appendix:passage_selector}
Our passage selection model is based on RoBERTa$_{\text{BASE}}$ \cite{liu_roberta_2019}. We feed each passage we select into the model as input and use an MLP on top of RoBERTa’s [CLS] token to predict if it is positive or negative. We then use this model to perform inference and obtain a score for every single passage in the whole wikipedia corpus. The top $N$ passages ranked by its score are served as the candidate pool we generate answers from. The model is optimized for a higher recall, such that the positive passages should be identified with a high probability. Our model achieved 84.7\% recall on the Natural Questions dev set.

\subsection{Further Details on Question Quality}
\label{appendix:qgen_assessment}

For NQ, we find that the retrieved questions are paraphrases of the test questions in the majority of cases. The test questions in TriviaQA are mostly very specific, and whilst retrieved questions in PAQ contain less details, they still usually have correct answers.
To better evaluate the quality of the questions, we conduct human evaluation on 50 random sampled questions generated from the wikipedia passage pool. We have the following observations: 1) the majority (82\%) questions accurately capture the context of the answer in the passage, and contain relevant and informative details to locate the answer. 2) 16\% questions fail to understand the semantics of certain types of answers. The failure can be traced to: \emph{Mistaking extremely similar entities.} For example, given the sentence ``The Kerch Peninsula is ... located at the eastern end of the Crimean Peninsula'' and the answer ``the Crimean Peninsula'', the generated question is ``what is the eastern end of the kerch peninsula''; \emph{Generalisation to rare phrases.} The digit combinations appearing in passages mostly stand for date or timestamp. However, it is not applied to all the cases and the model fails to capture that. For example for ``under a 109–124 loss to the Milwaukee Bucks'', the question is generated as ``when did ... play for the toronto raptors''. 3) Very few (2\%) questions mismatch question type words (Wh-types) with answers when the answers are words rarely asked (e.g. first).

\subsection{Further details on System Size vs Accuracy}
\label{appendix:memory}
\begin{figure}[t]
\centering
  \includegraphics[width=0.48\textwidth]{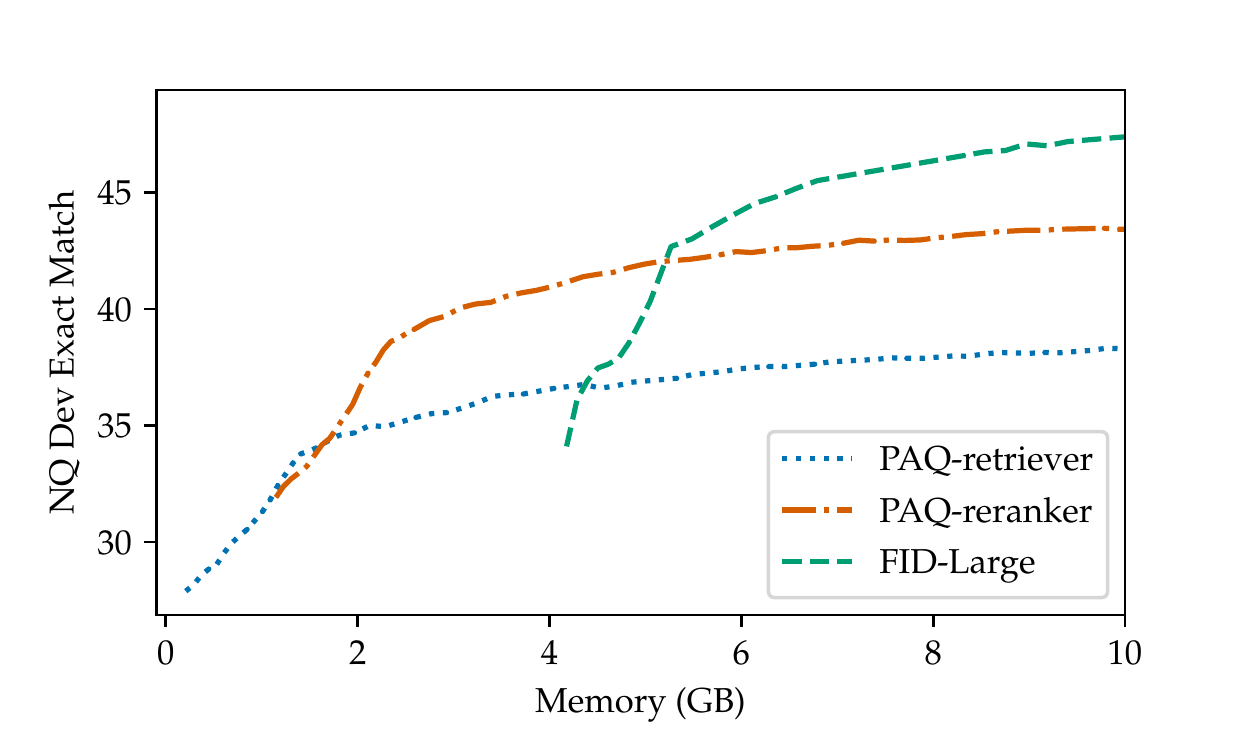}
  \caption{System size vs. accuracy for \QAEngine{}  and FID-large as a function of the number of items in the index, with different experimental setup than the main paper}
  \label{fig:memory_plot_different_setup}
\end{figure}

The system experiment described in \cref{sec:system_size_v_accuracy} measures the bytes required to store  the models, the text of the documents/QA-pairs, and a dense index.
For figure \ref{fig:memory_plot}, We assume models are stored at fp16 precision, the text has been compressed using LZMA\footnote{\url{https://tukaani.org/xz/}}, and the indexes both use 768d vectors, and Product Quantization. These are realistic defaults, with usage in the question answering literature~\cite{izacard_memory_2020,min_neurips_2020}.
The \QAEngine{} model used in this figure consists of an ALBERT-base retriever and ALBERT-xxlarge reranker, and the FID system consists of DPR~\cite{karpukhin_dense_2020} (itself consisting of two BERT-base retrievers) and a T5-large reader~\cite{raffel_exploring_2020}.
Using a different system setup (for example using full precision to store the models, no text compression, and fp16 index quantization, shown in figure \ref{fig:memory_plot_different_setup}) shifts the relative position of the curves in Figure \ref{fig:memory_plot}, but not the qualitative observation that \QAEngine{} models can be compressed to smaller sizes before significant drop in accuracy.

\subsection{Further details on Inference speed }
\label{appendix:speed}
The system used for inference speed benchmarking is a machine learning workstation with 80 CPU cores, 512GB of CPU RAM and access to one 32GB NVIDIA V100 GPU.
Inference is carried out at mixed precision for all systems, and questions are allowed to be answered in parallel.  Models are implemented in Pytorch~\cite{paszke_pytorch_2019} using Transformers~\cite{wolf_huggingfaces_2020}. Measurements are repeated 3 times and the mean time is reported, rounded to an appropriate significant figure. The HNSW index used in this experiment indexes all 65M PAQ QA-pairs with vectors of dimension 768 dimensions, uses an \texttt{ef\_construction} of 80, \texttt{ef\_search} of 32, and \texttt{store\_n} of 256, and performs up to 2048 index searches in parallel. This index occupies 220GB, but could be considerably compressed with techniques including scalar or product quantization, or training retrievers with smaller projection dimensions. The reader is referred to \citet{izacard_memory_2020} and \citet[][Appendix C]{lewis_retrieval-augmented_2020} for experiments demonstrating index compression with little-to-no drop in accuracy.

\subsection{Further details on Selective Question Answering}
\label{appendix:selective_qa}
\begin{figure}[t]
\centering
  \includegraphics[width=0.48\textwidth]{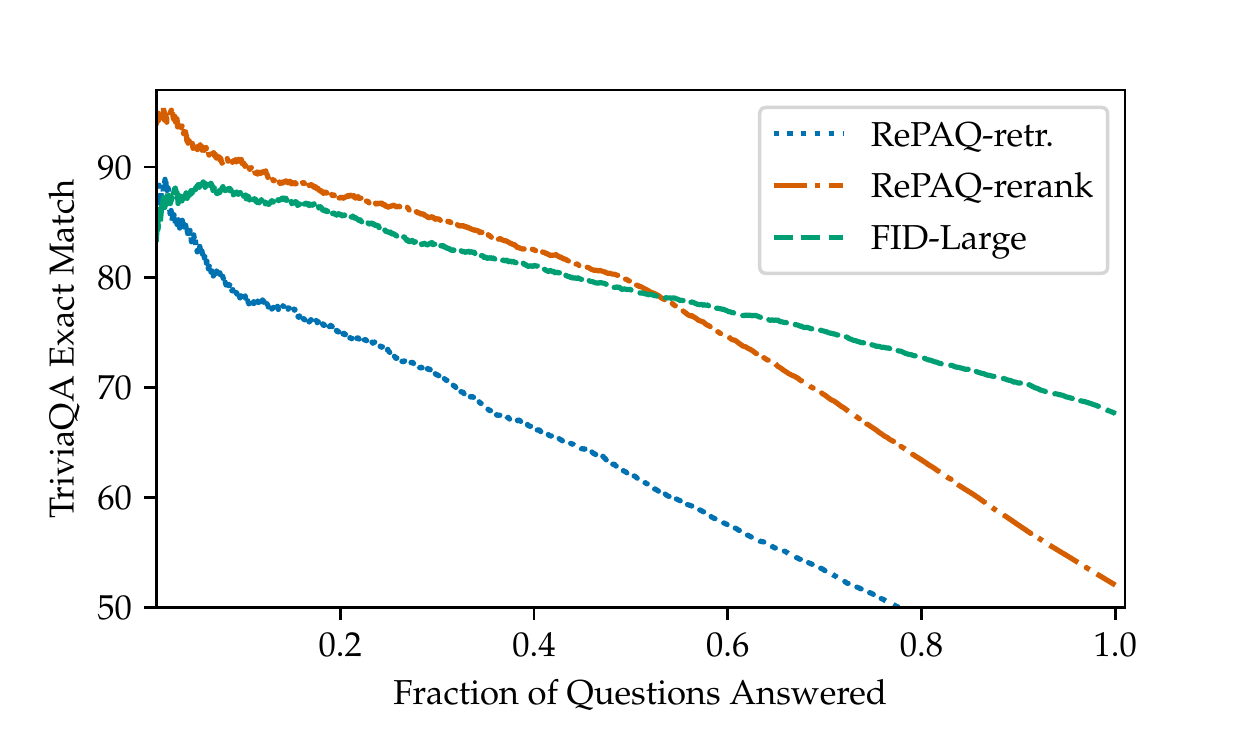}
  \caption{Risk Coverage Plot for FiD and \QAEngine{} on TriviaQA. FID has higher overall accuracy, but \QAEngine{} with a reranker still outperforms it for coverages $<$50\% }
  \label{fig:risk_coverage_tqa}
\end{figure}

Figure \ref{fig:risk_coverage_tqa} shows the Risk-Coverage plot for TriviaQA. The results are qualitatively similar as those for NQ (see Figure \ref{fig:risk_plot} in the main paper), although FiD's stronger overall performance on TriviaQA shifts its risk-coverage curve up the accuracy axis relative to \QAEngine{}. FiD also seems a little better calibrated on TriviaQA  than NQ, indicated by greater gradient. However, \QAEngine{} remains better calibrated than FiD, and outperforms it for answer coverages below 50\%.

\begin{figure}[t]
\centering
  \includegraphics[width=0.48\textwidth]{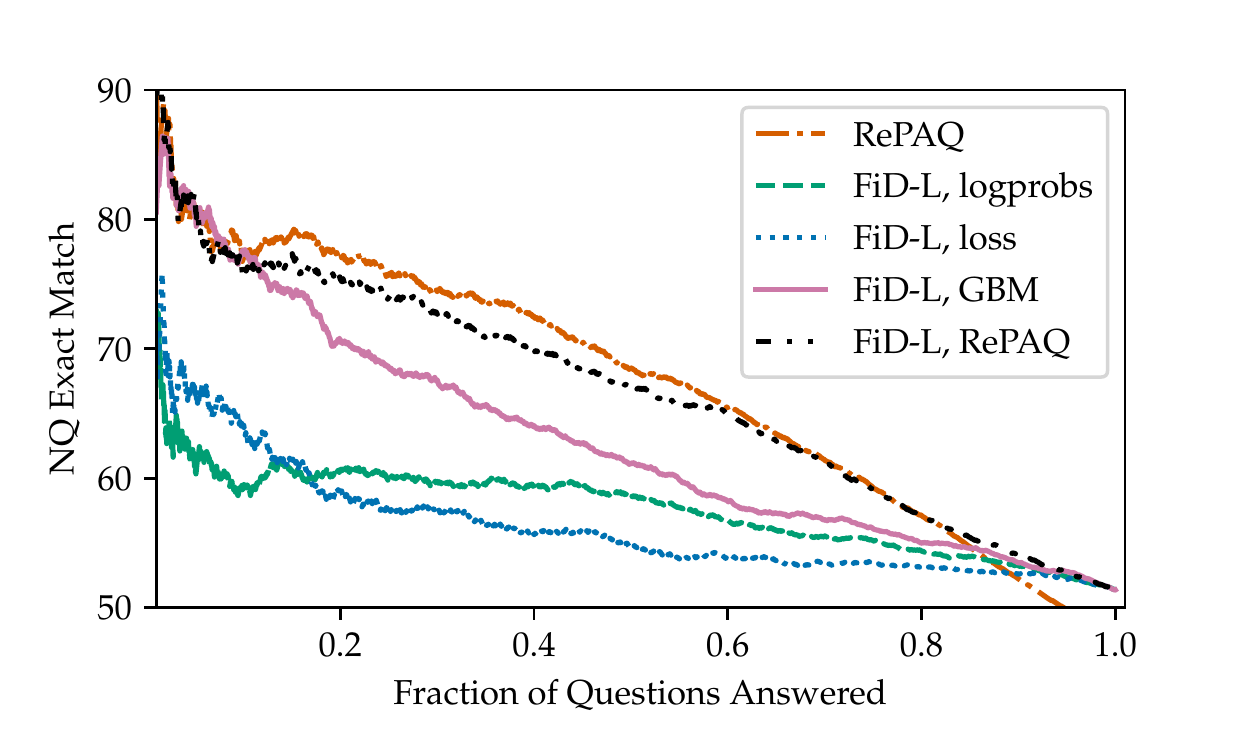}
  \caption{Risk Coverage Plot for different Calibration scores for FiD-Large (\QAEngine{} included for comparison). Using \QAEngine{}'s answer confidence scores to calibrate FiD leads to the best results for FiD}
  \label{fig:risk_coverage_metrics}
\end{figure}

We also investigate ways to improve the calibration of FiD-large on NQ, using post-hoc calibration, using an approach similar to \citet{jiang_how_2020}. We train a Gradient Boosting Machine \cite[GBM,][]{friedman_greedy_2001} on the dev set of NQ, which attempts to predict whether FID-large has answered correctly or not. The GBM is given FID's answer loss, answer log probability and the retrieval score of the top 100 retrieved documents from DPR. Figure \ref{fig:risk_coverage_metrics} shows these results. 
We first note that FiD-Large's answer loss and answer log probabilities perform similarly, and both struggle to calibrate FiD, as mentioned in the main paper. The GBM does improve calibration, especially at lower coverages, but still lags behind \QAEngine{} by 7\% EM at 50\% coverage.
We also note a surprising finding that we can acutally use \QAEngine{}'s scores  to calibrate FiD. Here, we use FiD's predicted answer, but \QAEngine{}'s confidence score to decide whether to answer or not. This result is also plotted in Figure \ref{fig:risk_coverage_metrics}, and results in the best risk-coverage curve for FiD. However, this is still not as well-calibrated as simply using \QAEngine{}, highlighting the strength of \QAEngine{} for selective QA further.

\subsection{Additional Model training details}
\QAEngine{} models were trained for up to 3 days on a machine with 8 NVIDIA 32GB V100 GPUs. Validation Exact match score was used to determine when to stop training in all cases. \QAEngine{} retrievers were trained using Fairseq~\cite{ott_fairseq_2019}, and rerankers were trained in Transformers~\cite{wolf_huggingfaces_2020} in Pytorch. 
The \KB{} BART CBQA models were trained in Fairseq for 6 days on 8 V100s, after which validation accuracy had plateaued. Standard BART hyperparameters were used, apart from batch size and learning rate, which were tuned to try to promote faster learning, but learning became unstable with learning rates greater than 0.0001. 
\end{document}